\documentclass[letterpaper, 10 pt, conference]{ieeeconf}
\IEEEoverridecommandlockouts
\pdfoutput=1 
\usepackage{graphics}
\usepackage{graphicx}
\usepackage{amsmath}
\usepackage{amssymb}
\usepackage{mathtools}
\usepackage{url}
\usepackage{booktabs}
\usepackage[font=small]{caption}
\usepackage{multirow}
\usepackage{afterpage}
\usepackage{url}
\usepackage{color}
\usepackage{threeparttable}
\usepackage{algorithm}
\usepackage{algorithmic}
\usepackage[algo2e]{algorithm2e}
\usepackage{bm}
\usepackage{latexsym}

\usepackage{color}
\usepackage{booktabs}
\usepackage{diagbox}
\usepackage{caption}
\usepackage{caption,subcaption}
\usepackage{array}

\definecolor{myRed}{rgb}{0.8, .2, .2}
\definecolor{myYellow}{RGB}{0, 0, 0}
\newcommand\revised[1]{\textcolor{myYellow}{#1}}

\newcommand{\ds}[1]{\textcolor{myYellow}{{#1}}}

\newcommand{\myEqRef}[1]{Eq.\ref{#1}}

\title{\LARGE \bf Single-Shot is Enough:  Panoramic Infrastructure Based Calibration of Multiple Cameras and 3D LiDARs}
\author{Chuan Fang$^{1}$, Shuai Ding, Zilong Dong, Honghua Li, Siyu Zhu, Ping Tan 
	\thanks{$^{1}$
		All the authors are with Alibaba Group, Hangzhou, China.
		{\tt\small fangchuan.fc@alibaba-inc.com}.}
}

\graphicspath{./figures/}

\begin{document}
	\maketitle 

\begin{abstract}
	
	The integration of multiple cameras and 3D LiDARs has become basic configuration of augmented reality devices, robotics, and autonomous vehicles. The calibration of multi-modal sensors is crucial for a system to properly function, but it remains tedious and impractical for mass production. Moreover, most devices require re-calibration after usage for certain period of time. In this paper, we propose a \textit{single-shot} solution for calibrating extrinsic transformations among multiple cameras and 3D LiDARs. We establish a panoramic infrastructure, in which a camera or LiDAR can be robustly localized using data from single frame. Experiments are conducted on three devices with different camera-LiDAR configurations, showing that our approach achieved comparable calibration accuracy with the state-of-the-art approaches but with much greater efficiency. 
	
\end{abstract}	

\section{Introduction}

The capability of sensing the surrounding environment is essential to augmented reality devices, robotics, and autonomous vehicles. 
These systems tend to integrate multiple cameras and LiDARs to overcome the limitation of individual sensors in terms of filed of view and functionality. 
Multi-modal sensors should be well calibrated before multifarious data can be precisely aligned and assembled together. 
Intrinsic calibration of cameras has been well studied in recent works~\cite{zhang2000flexible,kannala2006generic,mei2007single}. In contrast, extrinsic calibration of multi-model sensors remains a challenging task due to the variety of their spatial configuration and sensor types.

\begin{figure}[ht]
	\begin{subfigure}{0.16\textwidth}
		\includegraphics[width=\textwidth]{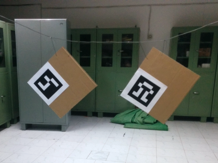}\hfill
		\caption{}
		\label{fig:cardboard}
	\end{subfigure}\hfil
	\begin{subfigure}{0.16\textwidth}
		\includegraphics[width=\textwidth]{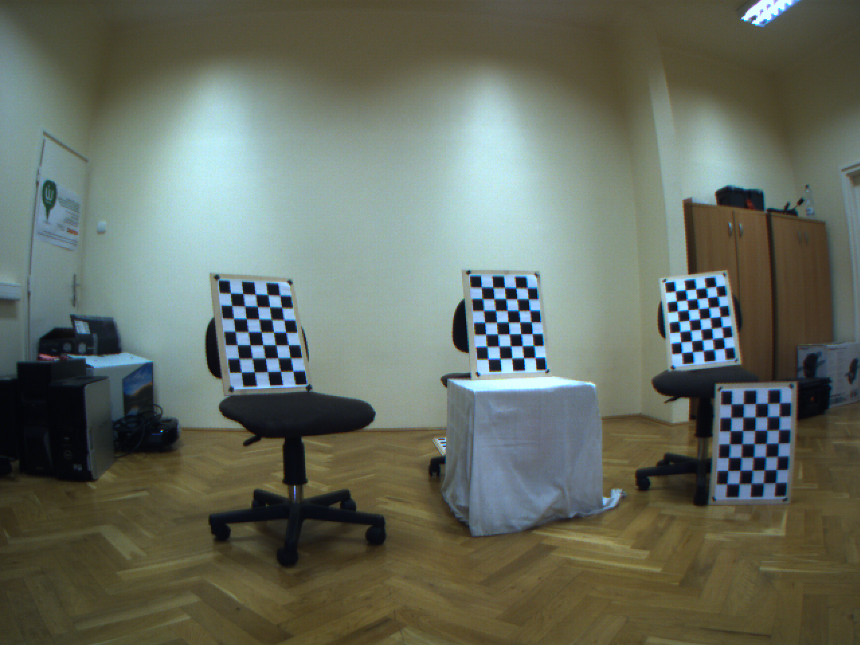}\hfill
		\caption{}
		\label{fig:multi-checkerboard}
	\end{subfigure}\hfil
	\begin{subfigure}{0.16\textwidth}
		\includegraphics[width=\textwidth]{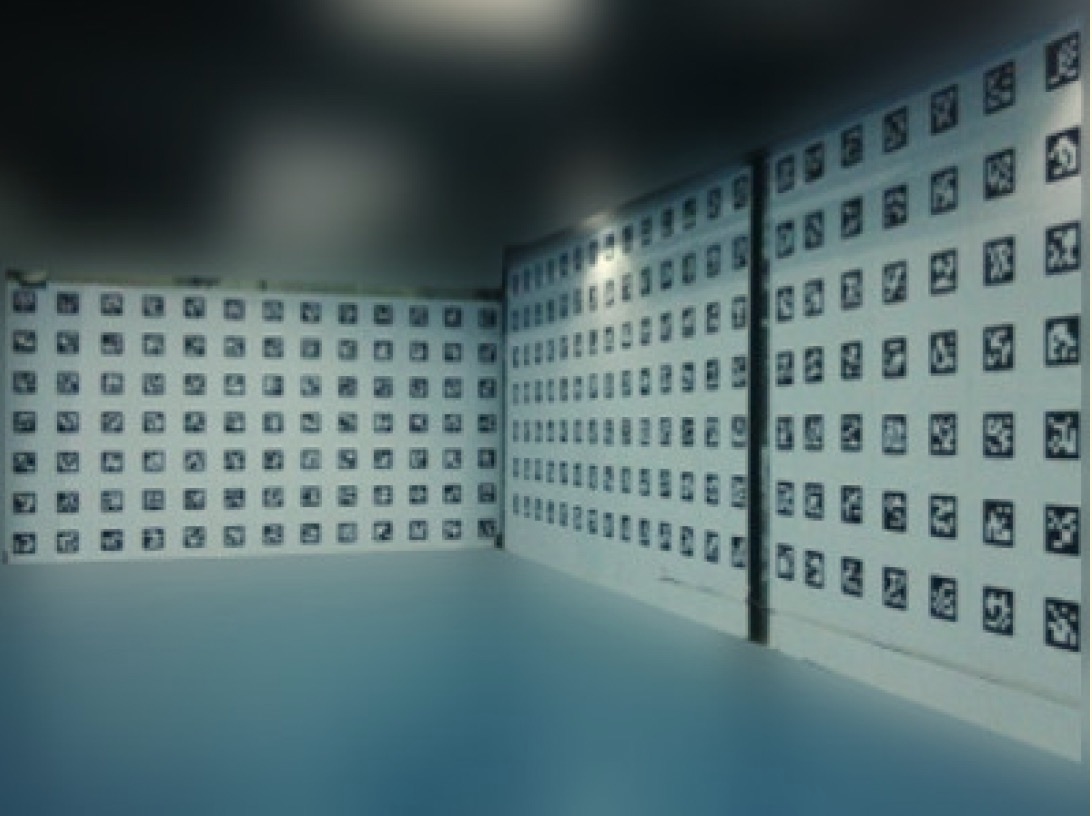}\hfill
		\caption{}
		\label{fig:aprilTag}
	\end{subfigure}\hfil
	\vspace{0.2cm}
	\begin{subfigure}{0.485\textwidth}
		\includegraphics[width=\textwidth]{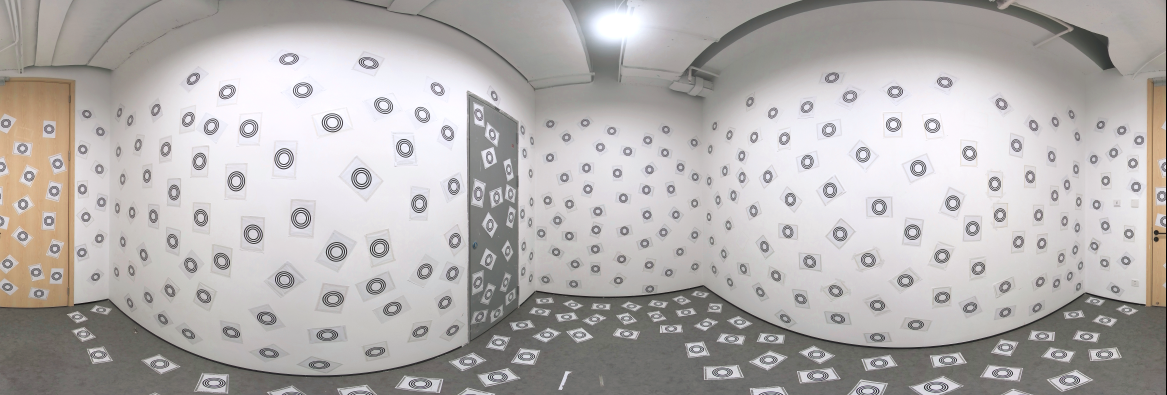}
		\caption{}
		\label{fig:panoramic}
	\end{subfigure}\hfil
	\caption{Various infrastructures with increasing scales. 
		(a) Cardboard based infrastructure~\cite{2017arXiv170509785D},
		(b) Checkerboard based infrastructure~\cite{pusztai2017accurate},
		(c) AprilTag based infrastructure~\cite{xie2018infrastructure}, 
		and 
		(d) Proposed panoramic infrastructure.}
	\label{fig:other_calib_ref_comparison}
\end{figure}

A straightforward solution to multi-sensor calibration is applying pairwise calibration methods and assembling resulted relative transformations. However, to calibrate pairwise camera-camera, camera-LiDAR or LiDAR-LiDAR extrinsics, separate pairwise calibration techniques are required, which possibly result in accumulated calibration errors and inefficiency. Fig.~\ref{fig:cardboard} presents two pieces of cardboard for camera-LiDAR calibration~\cite{2017arXiv170509785D}. 

In order to calibrate multiple sensors at the same time, large scale infrastructures are proposed by researchers~\cite{geiger2012automatic,pusztai2017accurate,xie2018infrastructure}. The key motivation is to match data from multiple sensors to a shared infrastructure, which should be large enough to cover the view of all sensors.
Fig.~\ref{fig:multi-checkerboard} shows the infrastructure in~\cite{pusztai2017accurate} with multiple checkerboards, while Fig.~\ref{fig:aprilTag} demonstrates that in~\cite{xie2018infrastructure} three planar walls covered by hundreds of AprilTags~\cite{olson2011apriltag} are involved. 
However, these two infrastructures are not panoramic. It means that only a portion of the system can be calibrated at the same time, therefore time-consuming multiple captures are inevitable. 

To overcome the limitations aforementioned, we propose a panoramic calibration infrastructure. As shown in Fig.~\ref{fig:panoramic}, such an infrastructure is a 720-degree room composed of piece-wise planar walls and floors, and it is covered by featureless fiducial markers~\cite{calvet2016detection}. 
\revised{We demonstrate that the featureless fiducial marker slightly outperforms Apriltag in terms of calibration accuracy, as discussed in Sec.~\ref{sec:exp_marker_comparison}}.
Notably, our infrastructure has nice characteristics as following:
\begin{enumerate}
	\item \textit{Panoramic}. The 720-degree setting naturally overlaps with the vision of all sensors in a multi-modal system, so that only a single capture from a single pose is adequate for localizing all sensors to the infrastructure. 
	\item \textit{Robust}. The randomly distributed featureless fiducial markers and the straight lines and corners from the piece-wise planar walls constitute salient features for robust pose estimation of cameras and LiDARs respectively.
	\item \textit{Unified}. The calibration of camera-camera, camera-LiDAR, and LiDAR-LiDAR configurations are all supported in this setting.
\end{enumerate}

To reconstruct the 3D structures of the infrastructure, we use a low-end stereo camera to capture its appearance and adopt a standard Structure-from-Motion (SfM) pipeline for sparse reconstruction.
We demonstrate that such a low-cost pipeline is able to achieve comparable reconstruction accuracy compared with the high-end scanner adopted in the work~\cite{xie2018infrastructure}.
Subsequently, we are able to efficiently calibrate extrinsic parameters of systems with multiple cameras and LiDARs in a single shot and \revised{achieve} comparable calibration accuracy with the state-of-the-art approaches.

In the experimental section, we verify the accuracy and efficiency of the proposed calibration method by multiple devices with different sensor configurations.  
Given a single shot of the to be calibrated device, our proposed method can even handle the complex and challenging cases in which the sensors do not share overlapping regions. 
Both quantitative and qualitative results show that the accuracy of our method is comparable to the mainstream approaches while the calibration efficiency is remarkably improved.

\section{Related work}
\label{sec:related}
In this section, we mainly discuss pairwise sensor calibration, and two categories of multi-sensor calibration, non-panoramic or panoramic. 

\subsection{Pairwise calibration}
The calibration of pairwise sensors, including camera-camera~\cite{zhang2000flexible}, camera-LiDAR~\cite{2017arXiv170509785D,zhou2012extrinsic,verma2019automatic} and LiDAR-LiDAR~\cite{jeong2018complex,gong2017target}, has been extensively studied in past decades.
Checkerboard has become popular reference object for camera-camera calibration since the pioneer work by Zhang et al.~\cite{zhang2000flexible}. 
Checkerboard is widely used for camera-LiDAR calibration~\cite{2017arXiv170509785D,zhou2012extrinsic,verma2019automatic} as well, where extra geometric features, e.g., borders and corners, are extracted to build correspondences between camera frames and LiDAR scans.  
LiDAR-LiDAR calibration requires reference objects with rich geometric features, i.e. corners of building structures in~\cite{jeong2018complex}.

\subsection{Non-panoramic multi-sensor calibration}

For systems of multiple cameras and 3D LiDARs, the infrastructure must possess rich geometric features as well as visual features.
The infrastructures in~\cite{geiger2012automatic,pusztai2017accurate} are composed of multiple checkerboards which provide salient geometric features on borders and corners, as well as visual features. Their infrastructures are not pre-reconstructed, therefore the views of sensors must overlap, which limits their practical usage. For example, a device containing multiple sensors without overlapping views cannot be calibrated using their infrastructures.

Xie~et~al.~\cite{xie2018infrastructure} pastes Apriltags~\cite{olson2011apriltag} on three planar walls, such infrastructure
does not require the overlapped region between any two sensors. The localization of cameras rely on Apriltags, while that of LiDARs rely on ICP~\cite{segal2009generalized}. 
But this approach still need multiple captures of each sensor to ensure that all sensors can be localized in the infrastructure.

Considering multi-camera configuration, Li~et~al.~\cite{li2013multiple} proposes a dedicated pattern to minimize overlapping area among neighboring cameras. Kumar~et~al.~\cite{kumar2008simple} uses a mirror to solve the extrinsics of non-overlapping multi-camera system, but the mirror has to be in the camera’s field of view and the entire pattern is visible in the camera. 
Lee~et~al.~\cite{leeunified} use a single checkerboard to solve extrinsic calibration problem of the multi-camera multi-LiDAR system, it defines local frames or global frames captured by either one or more sensors, which indicates it fails to deal with non-overlapping situations. 
\revised{ Owens~et~al.~\cite{owens2015msg} leverage the pairwise calibration combined with a graph-based global refinement to obtain the external parameters among all sensors. The same as \cite{leeunified}, the work in \cite{owens2015msg} fails to deal with non-overlapping situations.}

\subsection{Panoramic multi-sensor calibration}
Panoramic infrastructures are also proposed in literature for multi-sensor calibration, but they only calibrate systems with single type of sensors, either a multi-camera rig~\cite{heng2013camodocal} or a multi-LiDAR configuration~\cite{jiao2019automatic}.
In contrast, our panoramic infrastructure is suitable for calibrating multi-sensor systems with mixed sensor types.

\begin{figure*}[ht]
	\centering
	\includegraphics[scale=0.55]{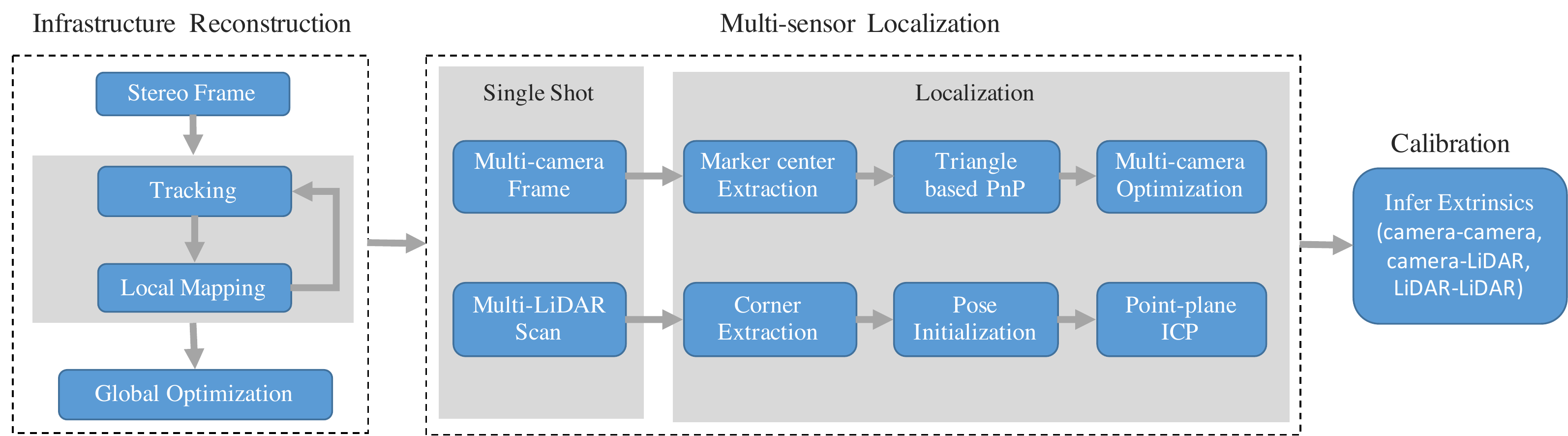}
	\caption{The framework of the proposed panoramic infrastructure based calibration of multiple cameras and 3D LiDARs.}
	\label{fig:system_framework}
\end{figure*}

\section{System overview}
\label{sec:overview}
The framework of the proposed multi-sensor calibration method based on panoramic infrastructure is shown in Fig.~\ref{fig:system_framework}.
In the process of infrastructure reconstruction, a standard incremental stereo SfM pipeline~\cite{SfM} described in Sec.~\ref{sec:methodology_map} is applied to obtain a sparse point cloud of the panoramic infrastructure.
Such a sparse reconstruction works as the calibration reference of the subsequent calibration process.
Based on the calibration reference, only a single scan of the sensor data including the multiple cameras and LiDARs is needed for the localization based calibration.
More specifically, camera poses are localized in the sparse map by visually matching the featureless fiducial markers, and a joint optimization of multiple cameras is applied subsequently (Sec.~\ref{sec:methodology_camera}).
Likewise, we localize the poses of LiDARs by the geometric features such as lines and points with respect to the same sparse reconstruction (Sec.~\ref{sec:methodology_LiDAR}).
Finally, we can derive the relative poses between arbitrary two sensors from their corresponding rigid body transformation with respect to the calibration reference.

In the following paragraphs, we define the notations and frame definitions adopted throughout the paper.
Note that all the sensors to be calibrated are rigidly connected.
We consider $(\cdot)^{W}$ as the world frame, $(\cdot)^{C}$ as the camera frame, and $(\cdot)^{L}$ as the LiDAR frame.
Given a device consisted of $N$ cameras and $K$ LiDARs, we regard the first camera $C_0$ as the reference of the camera frame and the first LiDAR $L_0$ as the reference of the LiDAR frame.
We use rotation matrix $R$ to represent rotation and 3D vector $t$ to denote the translation. $R^x_y$ and $t^x_y$ are respectively rotation and translation from $y$ frame to $x$ frame.

Moreover, we use $\{P\}$ to represent the 3D centers of featureless fiducial markers in the world frame, and $\{p\}$ the corresponding 2D centers of the featureless fiducial markers in the camera frame. Given that the intrinsics of the pinhole camera $C_i$ is defined as $K_i$, the projection function 
$\bm{\pi_{C}}$ can be formulated as:
\begin{equation}
	\begin{split}
		p_i \sim \bm{\pi_{C}}(K_i, & P, T^{C_i}_W) = K_i \cdot (R^{C_i}_W \cdot P + t^{C_i}_W) \\
		i &\in \{0,1,...,N-1\} 
	\end{split}
	\label{eq:eq_cam_project_func}
\end{equation}

Based on the notations defined above, the extrinsic calibration problem is to estimate the transformation $T^x_y$ between multiple sensors within the device, that is
\begin{equation}
	\begin{split}
		T^x_y &= \begin{bmatrix} R^x_y & t^x_y \\ 0 & 1 \end{bmatrix}  \\ 
		x,y \in \{L_0,&..., L_{K-1}, ...., C_0,...,C_{N-1}\}, x \neq y.
	\end{split}
	\label{eq:eq_extrinsic_def}	    
\end{equation}

\section{Methodology}
\label{sec:methodology}
In this section, we present the detailed description of the whole method, including the elaborate mechanism of panoramic infrastructure reconstruction, visual localization of multiple cameras, and LiDAR localization.

\subsection{Panoramic infrastructure reconstruction}
\label{sec:methodology_map}
We choose the circular marker as the featureless fiducial marker because the center detection of circular markers is generally acknowledged to be of high accuracy~\cite{calvet2016detection},~\cite{circular2000},~\cite{Accurate2009},\ds{~\cite{MALLON2007921}}.
\revised{The quantitative experiments shown in Sec.~\ref{sec:exp_marker_comparison} also demonstrate that the circular markers outperform the AprilTags in calibration accuracy.}
To reconstruct the panoramic infrastructure, a stereo SfM pipeline based on the featureless fiducial markers is proposed, and such a pipeline is composed of the following three steps: stereo frame tracking, local mapping and global optimization.

\begin{algorithm}[b]
	\caption{Triangle based stereo frame matching}
	\SetAlgoNoLine
	\KwIn{Two 3D point sets of consecutive stereo frames.}
	\KwOut{The relative pose $T_{rel}$ between consecutive stereo frames.}
	
	Enumerate all triangles in the 3D points denoted as $\mathcal{J}_{k}$ and $\mathcal{J}_{k-1}$ respectively.
	
	\For{each triangle in $\mathcal{J}_{k}$}{
		
		Find the corresponding triangle in $\mathcal{J}_{k-1}$ with the closest edge lengths, as the two red triangles in Fig.~\ref{fig:method_map_3dtrian_match},
		
		Estimate the rigid transform $T$ between the matched triangles with the Least-Squares Fitting method~\cite{LeastSquares1987},
		
		Calculate the inliers of $T$ similar to RANSAC;
	}
	Select the best rigid transform $T_{rel}$ with the most inliers.
	
	\algorithmicreturn~$T_{rel}$.
	\label{algo:algo_stereo_tra}
\end{algorithm}

\subsubsection{Stereo Frame Tracking}
In order to capture consecutive stereo frames, we move the stereo camera smoothly and capture the 720-degree room in a uniform pattern.
Analogous to the existing stereo SfM methods~\cite{SfM}, the featureless circle center of both left and right frames are detected by the method~\cite{calvet2016detection}.
Given that both the intrinsics and exstrinsics of the stereo camera are calibrated, we obtain the feature correspondences across the stereo cameras by the standard epipolar search.
Once the feature correspondences between the left and right frames are established, the corresponding sparse 3D points can be triangulated.
To build the correspondences of sparse 3D points between consecutive stereo frames, we propose the following triangle based matching algorithm and provide the details in Algorithm~\ref{algo:algo_stereo_tra}.

To increase the density of the infrastructure reconstruction, the unmatched 3D points of a specific stereo frame still remained in the final reconstruction.

\begin{figure}
	\centering
	\begin{subfigure}{0.252\textwidth}
		\centering
		\includegraphics[width=\textwidth]{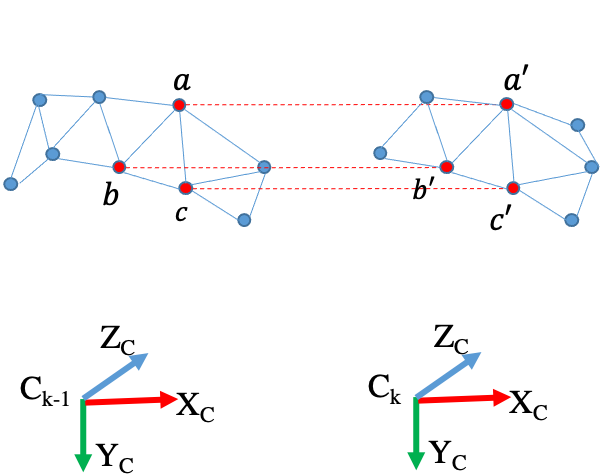}
		\caption{}
		\label{fig:method_map_3dtrian_match}
	\end{subfigure}\hfil
	\hspace{1em}
	\begin{subfigure}{0.19\textwidth}
		\centering
		\includegraphics[width=\textwidth]{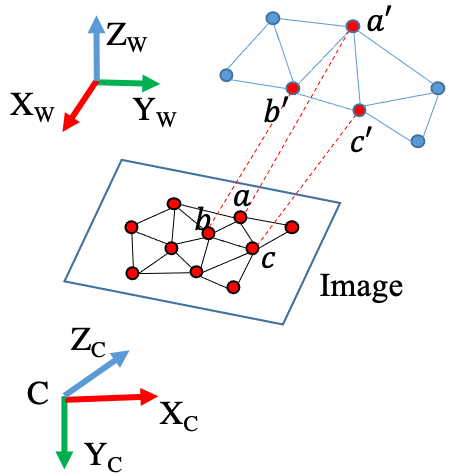}\hfill
		\caption{}
		\label{fig:method_vl_2dtrian_match}
	\end{subfigure}\hfil
	\caption{The visual demonstration of 3D-3D and 2D-3D feature correspondence matching by triangle matching. (a) The 3D-3D triangle matching in stereo frame tracking. (b) The 2D-3D triangle matching in camera localization.}
	\label{fig:triangles_matching}
\end{figure}

\subsubsection{Local Mapping}
After each operation of stereo frame tracking, a loop closure detection is proposed to obtain more robust 3D correspondences between the current and previous stereo frames.
The strategy of the proposed loop closure detection is to project the sparse points of the current stereo frame to the other stereo frames which are far away from the current stereo frames in terms of time sequence.
Then, the stereo frame which has the most co-visible sparse points with the current stereo frame is retrieved, and their matching correspondences are obtained by performing Algorithm~\ref{algo:algo_stereo_tra}.

\subsubsection{Global optimization}
Since the incrementally reconstructed result suffers from the accumulated errors, a Bundle Adjustment (BA) optimization is applied to refine all the camera poses and sparse points simultaneously.
Before that, we merge the nearby sparse points with the distance less than $2 cm$ into one.
Afterwards, a global BA is followed to minimize the following reprojection error function:
\begin{equation}
	\sum_{k}\sum_{i}\bm{e}_r = \left\| \bm{z}_{i,k} - \bm{\pi_C}(K_i, P^W_i, T^W_k) \right \|,
	\label{eq:eq_stereo_sfm_ba}
\end{equation}
where $k$ denotes the stereo frame index, $i$ denotes the index of the sparse point visible in the $k^{th}$ stereo frame, and $\bm{z}_{i,k}$ represents the projection of $P^W_i$ in $k^{th}$ stereo frame.

\subsection{Camera localization}
\label{sec:methodology_camera}
Once the sparse map of the panoramic infrastructure is available, we can accurately localize pre-calibrated cameras in only single shot. Firstly, the localization of each camera in the calibration infrastructure provides the initial extrinsic between multi-camera. Then we apply the geometric constrains of multiple cameras to further refine the extrinsic parameters by an iterative Levenberg–Marquardt (LM) algorithm~\cite{more1978levenberg}. 

Since the fiducial markers in the calibration reference are featureless, the traditional visual localization methods~\cite{piasco2018survey} fail to estimate the poses of the multiple cameras to be calibrated.
As shown in Fig.~\ref{fig:method_vl_2dtrian_match}, similar to the triangle based 3D point matching approaches introduced in stereo frame tracking, we can overcome the absence of salient texture based features of both images and 3D sparse points and apply the triangle based PnP combined with a RANSAC algorithm~\cite{fischler1981random} to obtain localized camera poses. In the following, we elaborate the implementations step by step.

\subsubsection{2D delaunay triangulation}
With the help of the marker detection algorithm~\cite{calvet2016detection}, we can have the result of marker centers of the current image denoted by $\mathbf{x_i}=\{p_0, p_1,..., p_k\}$.
Then, we apply the 2D delaunay triangulation in $\mathbf{x_i}$, and triangles with inner angle $\theta \leq 20^{\circ}$ and edge length ratio $r \geq 5$ are removed for a robust estimation.
Finally we can obtain the set of 2D triangles defined by:
\begin{equation}
	\begin{split}
		D_\mathbf{x_i} &= \{d_0, d_1, ..., d_j\}, \\
		& d_j =(p_a, p_b, p_c), \\
		a \neq b \neq c,\; & \text{and}\; a,b,c \in \{0,1,...,n-1\}.
	\end{split}
\end{equation}

\subsubsection{3D delaunay triangulation}
We propose a 3D delaunay method to associate 2D triangles $D_\mathbf{x_i}$ with 3D triangles $D_{P}$.
After enumerating all the triangle combinations of $\{P\}$, where the number of 3D points is $M$ in the infrastructure reconstruction, we can have a total number of $C_{M}^{3}$ 3D triangles. In order to speed up the calculation, 3D triangles with side length longer than $1m$ are discarded. 
Finally, we have a set of all possible 3D triangles, presented as:
\begin{equation}
	\begin{split}
		D_{P}=&\{D_0, D_1, ..., D_h, ..., D_{H-1}\},\\
		&D_h = (P_a, P_b, P_c), \\
		a \neq b \neq c,\;& \text{and}\; a,b,c \in \{0,1,...,H-1\}. 
	\end{split}
\end{equation}

\subsubsection{Triangle based PnP}
Given 2D triangle set $D_\mathbf{x_i}$ on $I_i$ and 3D triangle set $D_{P}$, the AP3P algorithm~\cite{ke2017efficient} is applied to obtain 2D-3D triangle matches. The AP3P algorithm is used in conjunction with RANSAC~\cite{fischler1981random} to remove outliers. The details of 2D-3D triangles based PnP algorithm are depicted in Algorithm~\ref{algo:algo_visual_local}.
\begin{algorithm}[H]
	\SetAlgoNoLine
	\caption{Triangle based PnP}
	\KwIn{ 2D triangle set $D_\mathbf{x_i}$, 3D traingles set $D_{P}$. }
	\KwOut{Camera pose $T^{C_i}_W$ }
	Randomly select $d_j \in D_\mathbf{x_i}$; 
	
	\For{ each triangle in $D_h \in D_P$}{
		$T^{C_{i,h}}_W$ $\leftarrow$ solvePnP($d_j$, $D_h$);
		
		Calculate inlier of $T^{C_{i,h}}_W$ similar to RANSAC;
	}
	
	Select best $T^{C_i}_W$;
	
	\algorithmicreturn~$T^{C_i}_W$
	\label{algo:algo_visual_local}
\end{algorithm}

\subsubsection{Multi-camera Optimization}
After the rough visual localization of each camera, we further optimize camera poses $\chi = \{T^{C_i}_W\}$ of $\{C_i\}$ by minimizing the following cost function :
\begin{equation}
	\underset{\chi}{argmin} {{\sum_{i \in N}} (\bm{\pi_{C_i}}(K_i, P^W_m, T^{C_i}_W) - p_i)^2}
	\label{eq:eq_cam_extrin_opt_2}
\end{equation}
\revised{Intuitively, the camera pose $T^{C_i}_W$ with respect to the panoramic infrastructure is refined by minimizing the reprojection errors of the 2D projections of the marker centers and their corresponding visual observations. 
	Given the camera poses obtained by the non-linear optimization defined in \myEqRef{eq:eq_cam_extrin_opt_2}, we can infer the extrinsics of any two cameras.}

\subsection{LiDAR localization}
\label{sec:methodology_LiDAR}
The localization of LiDAR mainly depends on geometrical information, such as spatial lines and corners in the point clouds. 
Here, the corner of the referenced reconstruction can be defined as a 4-tuple $(c^W, s^W_0, s^W_1, s^W_2)$, where $s^W_0, s^W_1, s^W_2$ denote the direction vectors of three corner lines and $c^W$ denotes the corner point in referenced panoramic infrastructure. Moreover, $(c^L, s^L_0, s^L_1, s^L_2)$ denotes the scene corner sampled by the LiDAR.
\revised{It should be noted that the spatial lines and corners are the intersections of the three wall planes which can be obtained by plane fitting. 
	And we make the three direction vectors orthogonal based on the approach proposed in~\cite{golub2013matrix}}.

The LiDAR pose with respect to the referenced panoramic infrastructure is denoted as $\begin{bmatrix} R^W_L t^W_L \end{bmatrix}$.
\revised{When the LiDAR is placed in the infrastructure, the rough orientation of the LiDAR is known, and we can manually assign the correspondences of spatial lines and corners between the LiDAR scan and the referenced panoramic infrastructure.}
After obtaining the correspondences of three spatial lines whose direction is different from each other and one spatial point between the referenced panoramic infrastructure and a LiDAR, the pose of a LiDAR scan with respect to the calibration reference can be computed by the following two equations:
\begin{equation}
	[s^W_0 , s^W_1 , s^W_2] = R^W_L[s^L_0,  s^L_1, s^L_2]
	\label{eq:eq_lidar_localize_R}
\end{equation}
\begin{equation}
	c^W = R^W_L \cdot c^L + t^W_L
	\label{eq:eq_lidar_localize_t}
\end{equation}	

The operation above is able to provide a rough pose of the input LiDAR scan. To further optimize the pose of the input LiDAR scan, we densify the sparse reconstruction of the panoramic infrastructure by plane fitting and perform the Iterative Closest Point (ICP) optimization algorithm~\cite{segal2009generalized} between the LiDAR scan and the densified reference reconstruction.

With respect to the referenced panoramic infrastructure, we finally obtain the localized poses of multiple cameras and LiDARs. It is straightforward to derive the relative poses between arbitrary two sensors.

	\section{Experimental results}
\label{sec:experiments}
\subsection{Experimental Setups}
\label{sec:exp_setup}
CCTag~\cite{calvet2016detection} is chosen to be the featureless fiducial marker used in our experiments, and we convert a storeroom \revised{(about 3m$\times$4m)} into a calibration infrastructure according to section  \ref{sec:methodology_map}. About 340 CCTags without ID information are printed on A4 papers and randomly pasted on the floor and four walls. The ceiling is ignored because five planes are enough for constraining the ICP procedure. \revised{The stereo camera used for reconstructing the calibration infrastructure is composed of two synchronized MV-CA050-20UM\footnote{https://en.hikrobotics.com/machinevision/visionproduct?typeId=78\&id=43}}.

\revised{To evaluate the efficiency of the proposed approach, we conduct all the experiments on a computer with Intel i7-8700k CPU@3.7GHz. As for the computational time of the main stages, the average localization time of one camera is \textbf{3.797} seconds, and the LiDAR localization costs \textbf{0.855} seconds, in which the manual intervention time is omitted.}

\begin{figure}
	\begin{subfigure}{0.24\textwidth}
		\includegraphics[width=\textwidth]{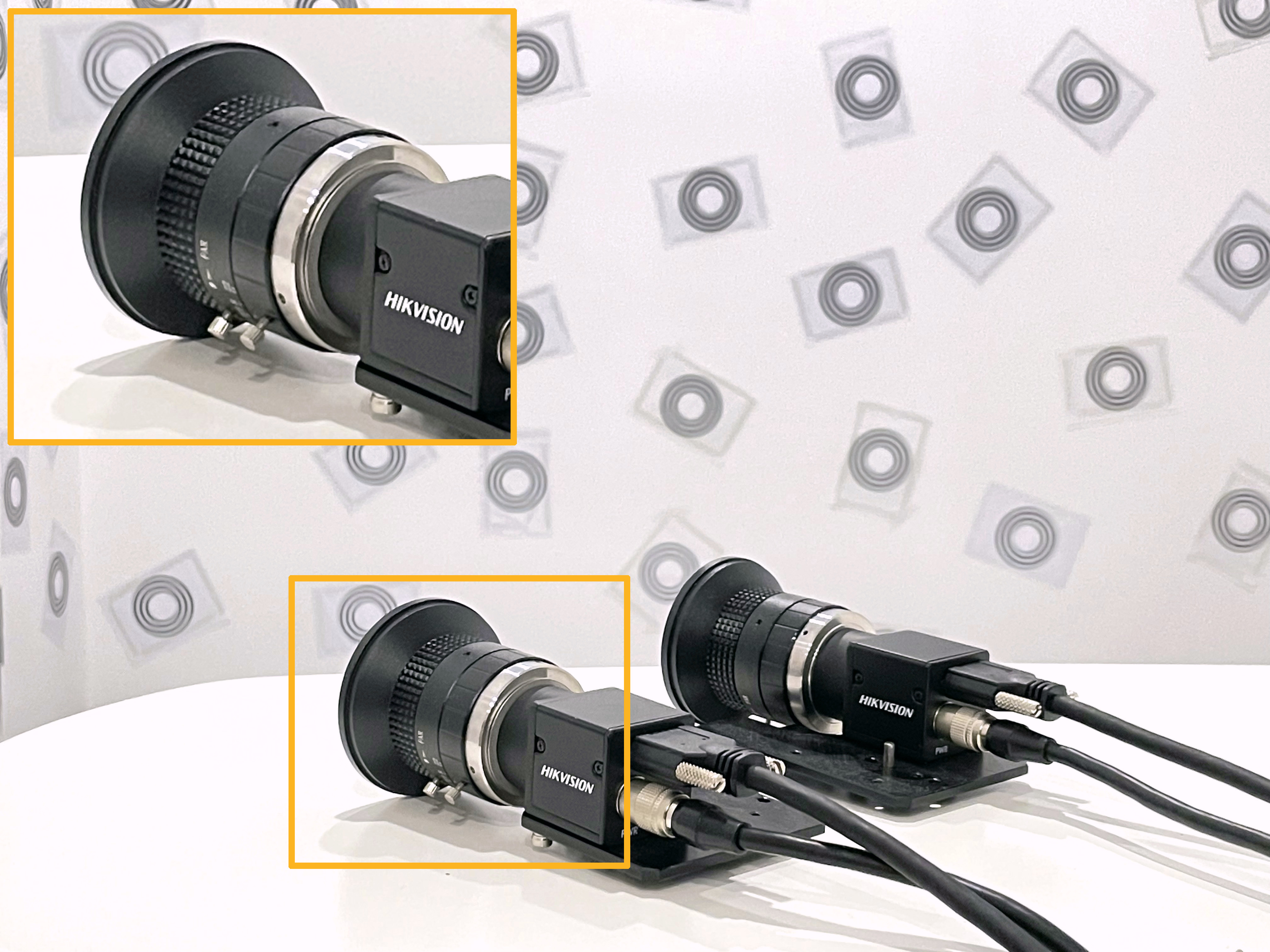}\hfill
		\caption{}
		\label{fig:hik}
	\end{subfigure}\hfil
	\begin{subfigure}{0.24\textwidth}
		\includegraphics[width=\textwidth, angle=0]{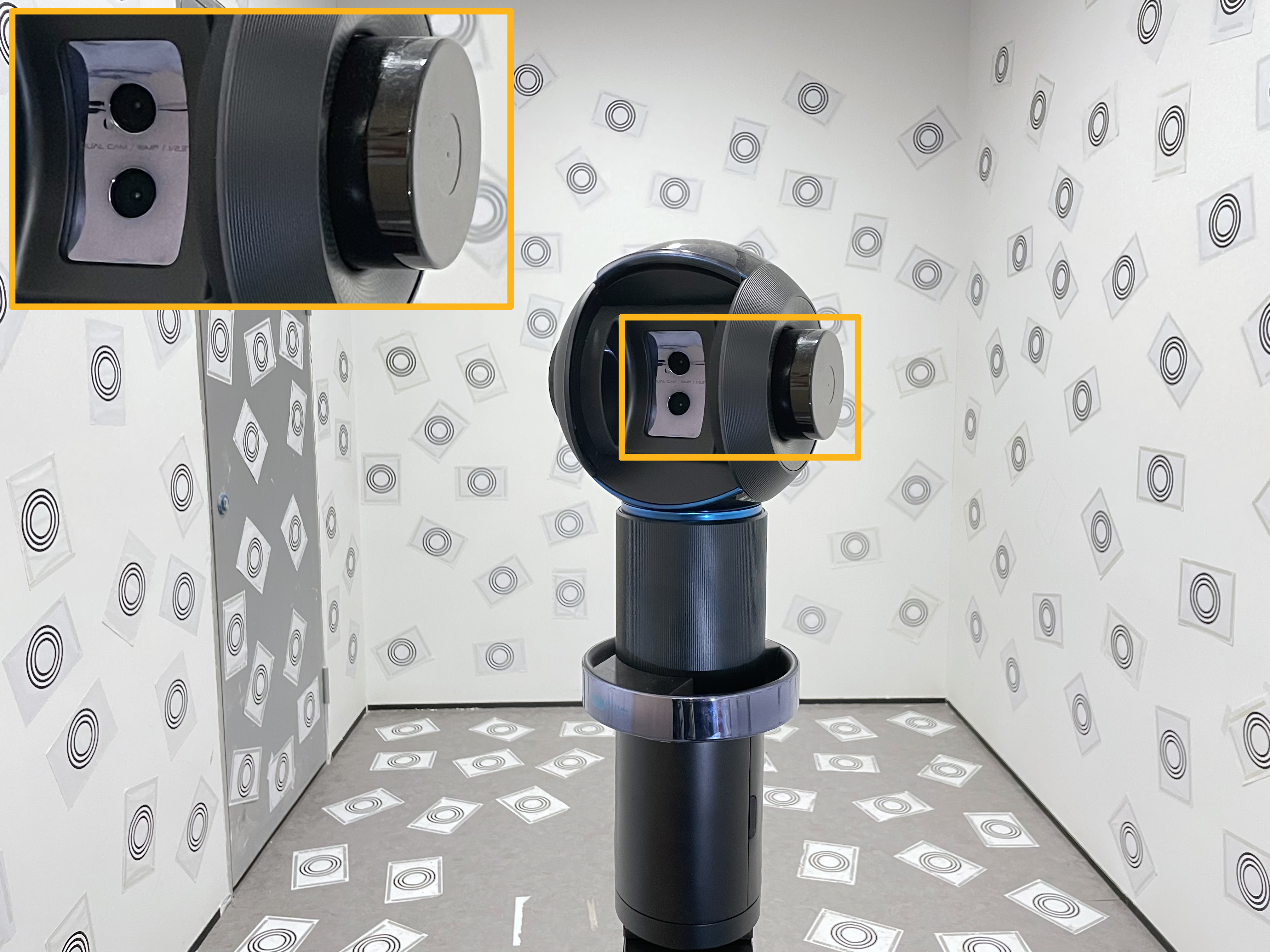}\hfill
		\caption{}
		\label{fig:kaleido}
	\end{subfigure}\hfil
	\vspace{0.2cm}
	\begin{subfigure}{0.24\textwidth}
		\includegraphics[width=\textwidth, angle=0]{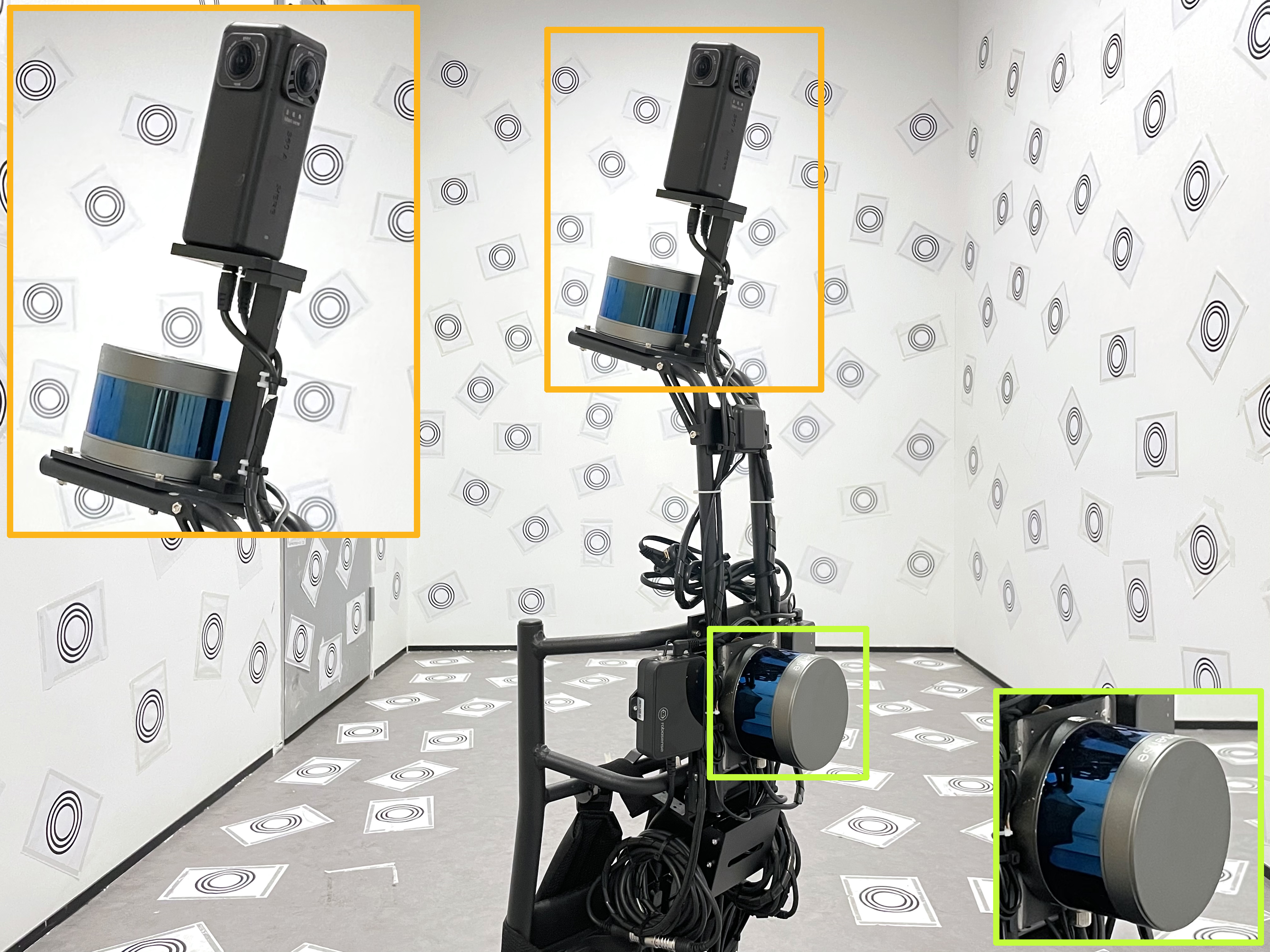}\hfill
		\caption{}
		\label{fig:backpack}
	\end{subfigure}\hfil
	\begin{subfigure}{0.24\textwidth}
		\includegraphics[width=\textwidth]{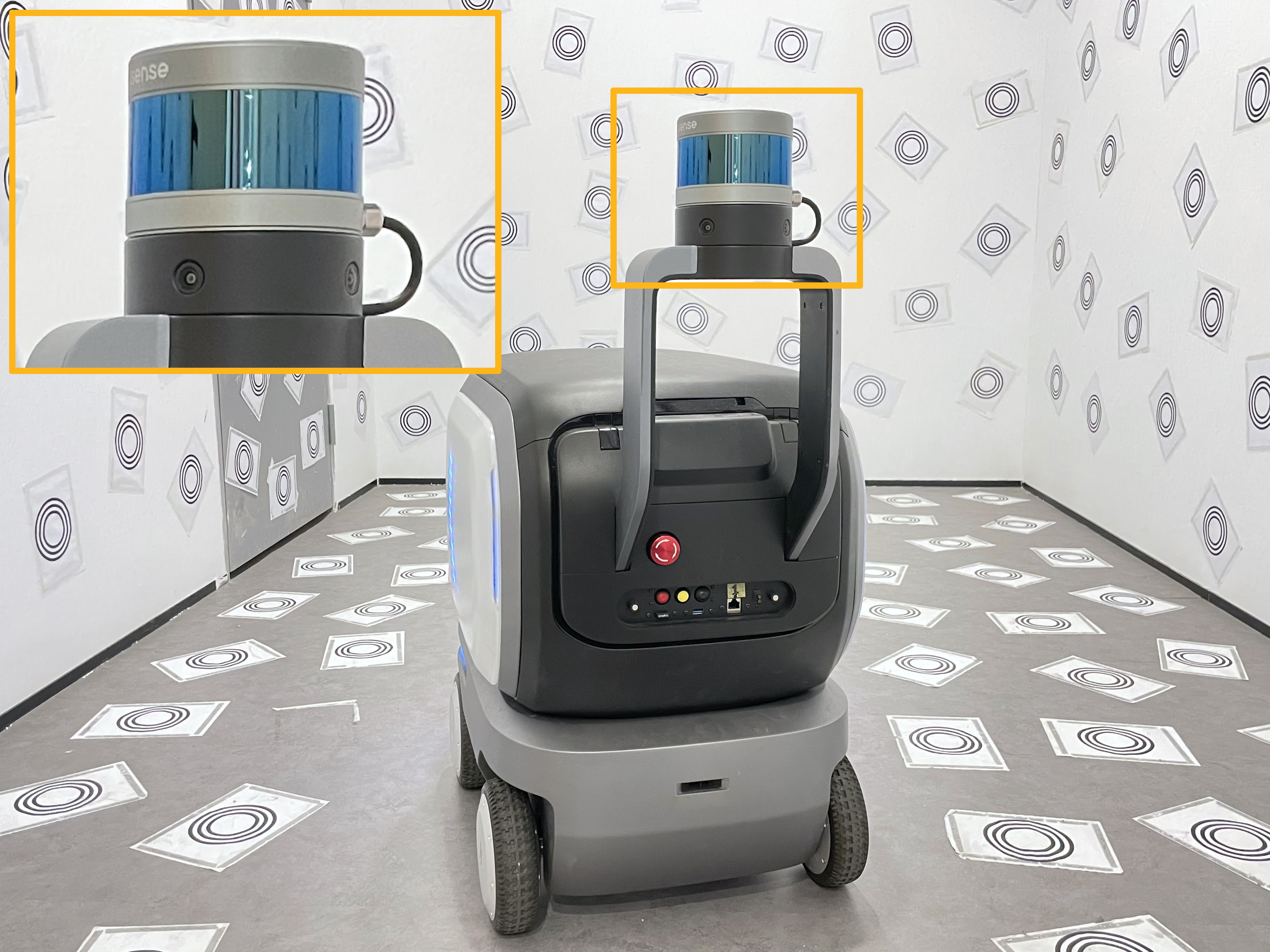}\hfill
		\caption{}
		\label{fig:mobile_robot}
	\end{subfigure}\hfil
	\caption{Devices used in the experiments with different sensors configurations highlighted: (a) The Stereo Camera Rig, (b) The Terrestrial Laser Scanner, (c) The Backpack Scanner, (d) The Mobile Robot.}
	\label{fig:hardwares}
\end{figure}

Four types of device are adopted in the experiments, as shown in Fig.~\ref{fig:hardwares}: 
\renewcommand{\labelenumi}{(\alph{enumi})}
\begin{enumerate}
	\item Stereo camera (resolution: 1280$\times$1024);
	\item Terrestrial Laser Scanner (TLS) composed of one 2D LiDAR and two cameras (resolution: 4608$\times$3456) mounted on a servo motor;
	\item Backpack scanner with two 3D LiDARs and four fisheye cameras (resolution: 2304$\times$3840);
	\item Mobile robot with one 3D LiDAR and four wide-angle cameras (resolution: 1280$\times$720).
\end{enumerate}

\subsection{Reconstruction accuracy}
\label{sec:exp_recons_accuracy}
\begin{figure}[t]
	\centering
	\begin{subfigure}{0.45\textwidth}
		\includegraphics[width=\textwidth]{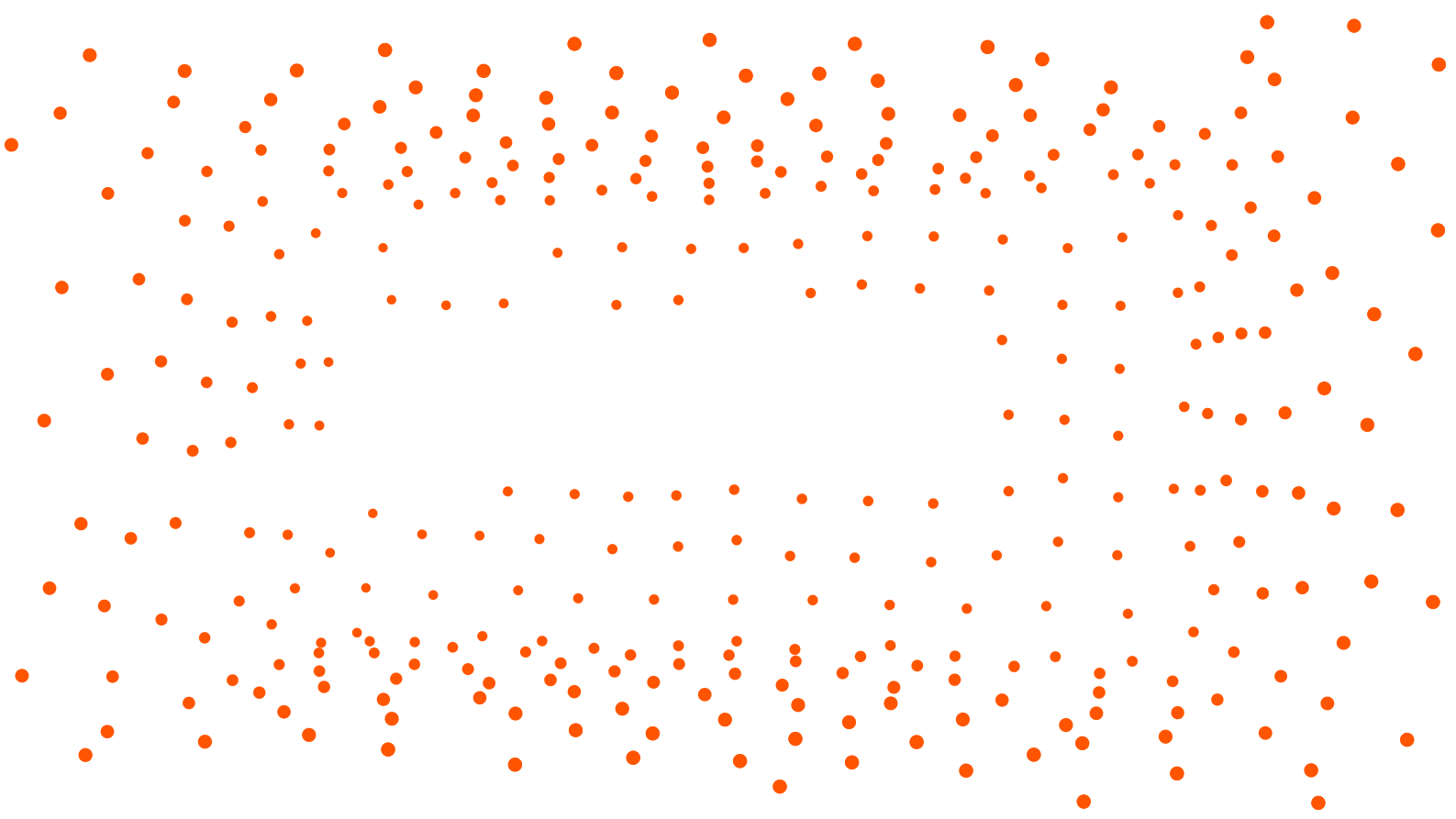}
		\caption{}
		\label{fig:sparse}
	\end{subfigure}\hfil
	\vspace{0.2cm}
	\begin{subfigure}{0.45\textwidth}
		\includegraphics[width=\textwidth]{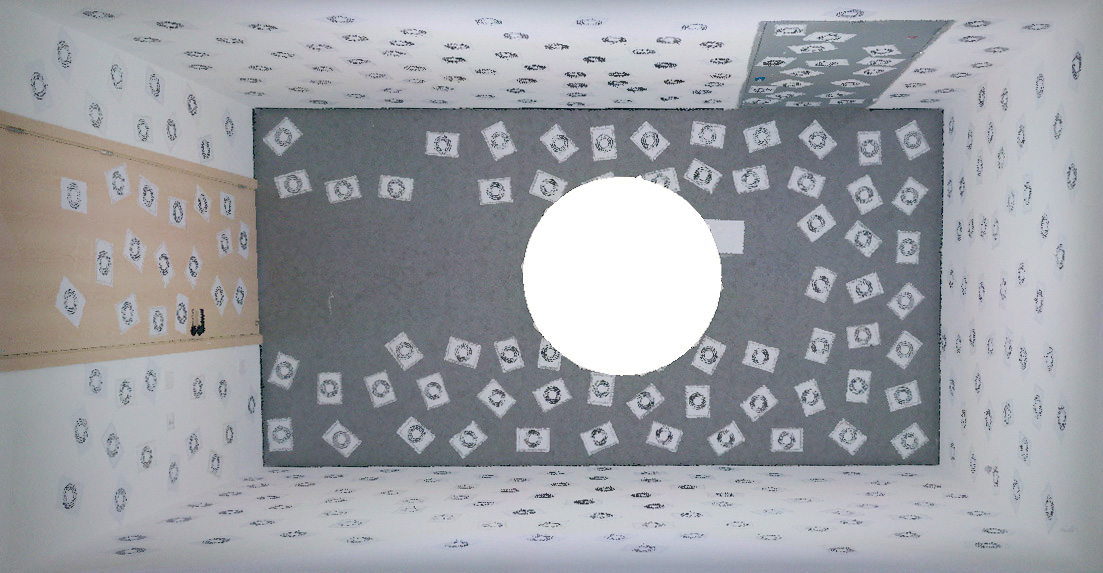}
		\caption{}
		\label{fig:Leica}
	\end{subfigure}\hfil
	\caption{The reconstruction of the storeroom: (a) The panoramic map points reconstructed by the proposed stereo SfM, (b) The dense point cloud reconstructed by the Leica BLK360. }
	\label{fig:room}
\end{figure}

The panoramic map points reconstructed by the proposed stereo SfM algorithm in Section~\ref{sec:methodology_map} are shown in Fig.~\ref{fig:sparse}. 
To verify the reconstruction accuracy, a high-end Leica BLK360\footnote{https://leica-geosystems.com/products/laser-scanners/scanners/blk360} is introduced to reconstruct the storeroom shown in Fig.~\ref{fig:Leica}, the captured dense point cloud is of high accuracy and can be regarded as the Ground Truth (GT). The reconstruction accuracy is evaluated by comparing the distance between the reconstructed sparse map points and the 3D centers of the CCTags from the reconstruction results of Leica BLK360, which are extracted by the following steps: \romannumeral1. Render the colored 3D points to an image view given a virtual camera pose inside the storeroom; \romannumeral2. Detect the 2D centers of the CCTags in the image view; \romannumeral3. Compute the ray vectors of the 2D centers and the intersections with the wall or the floor, which is considered as the 3D centers of the CCTags. 

It should be noted that the ICP~\cite{segal2009generalized} is utilized to align the two sparse map points before evaluation. 
\ds{Since the reconstruction accuracy heavily depends on the calibration of the stereo camera, we use Kalibr~\cite{kalibr} to accurately calibrate it. The pinhole camera model and the rad-tan distortion model are adopted. We calibrate the stereo camera and reconstruct the storeroom twice, and the corresponding results are presented in TABLE~\ref{tab:stereo_calib_result}. The results show that we can obtain a fairly accurate reconstruction result (about 0.1\%) based on a highly accurate calibration of the stereo camera.}

\begin{table}
	\centering
	\caption{\revised{Calibration results of the stereo camera used in reconstruction stage, and the corresponding reconstruction accuracy.}}
	\begin{tabular}{>{\centering\arraybackslash}m{0.5cm}  >{\centering\arraybackslash}m{1.5cm}  >{\centering\arraybackslash}m{1cm}  >{\centering\arraybackslash}m{2cm} >{\centering\arraybackslash}m{1.5cm}} 
		\toprule
		\revised{Times} & \revised{Focal (px) Left Right} &  \revised{Baseline (cm)} & \revised{Mean Reconstruction Error (cm)} & \revised{Mean Plane Fitting Error (cm)}\\ 
		\midrule
		\revised{1} & \revised{1746 1744} & \revised{12.09} & \revised{0.4(0.1\%)} & \revised{0.1}\\ 
		\revised{2} & \revised{1747 1744} & \revised{12.09} & \revised{0.5(0.12\%)} & \revised{0.11}\\ 
		
		\bottomrule
	\end{tabular}
	\label{tab:stereo_calib_result}
\end{table}

\subsection{Calibration accuracy}
\label{sec:exp_calib_accuracy}
Three experiments are conducted to evaluate the calibration accuracy of camera-camera, camera-LiDAR and LiDAR-LiDAR, respectively.
\revised{The structural measurements calculated from CAD drawings are treated as the GT.} To reduce randomness, all the calibration procedures are repeated five times, and the average results are presented. 
\revised{To evaluate how the reconstruction accuracy affects the calibration, we add different levels of Gaussian noises to the reconstruction results. We find that our method gets unacceptable calibration results when the noise standard deviation $\sigma > 1.5cm$(equals to 20.9mm reconstruction error). Specifically, $\mathcal N_0$ and $\mathcal N_1$ are used to represent Gaussian noise $\mathcal N(0,0.5cm)$ and $\mathcal N(0,1cm)$, respectively. The corresponding reconstruction errors of $\mathcal N_0$ and $\mathcal N_1$ are 9.9mm and 16mm, respectively.}
\subsubsection{camera-camera extrinsic calibration}
\label{sec:exp_cam_cam}
The extrinsic calibration between two cameras is performed on the stereo camera rig as shown in Fig.~\ref{fig:hik}. The baseline ($12cm$) provided by the manufacturer is considered as the Ground Truth. The calibration results of our proposed method are compared against the results of Kalibr toolbox~\cite{kalibr}, as presented in TABLE~\ref{tab:experiment_cam_cam_accu}. The results show that the proposed method using only one single frame achieves comparable performance with Kalibr. 

\begin{table}[t]
	\centering
	\caption{Camera-camera extrinsic calibration results.}
	\begin{tabular}{c c c c} 
		\toprule
		Method & Frames & Baseline (cm) & Reprojection Error (px)\\ 
		\midrule
		GT & - & 12  & -\\ 
		Kalibr~\cite{kalibr} & 45 & 12.022  & 0.13\\ 
		Our w/o noise & 1 & 11.984 & 0.47\\
		\revised{Our w $\mathcal N_0$}  & \revised{1} & \revised{11.918} & \revised{1.58}\\
		\revised{Our w $\mathcal N_1$}  & \revised{1} & \revised{12.947} & \revised{1.97}\\
		\bottomrule
	\end{tabular}
	\label{tab:experiment_cam_cam_accu}
\end{table}

\begin{table}[t]
	\centering
	\caption{Camera-LiDAR extrinsic calibration results. The Rotation is represented by Euler angles of the XYZ order, the Translation is expressed in the three directions of XYZ.}
	\begin{tabular}{c c c c} 
		\toprule
		Method & Scans & Rotation (deg) & Translation (cm)\\
		\midrule
		\revised{GT}  & \revised{-} & \revised{(90.00, -45.00, 0.00)} & \revised{(0.0, -5.93, -5.497)} \\
		Zhang~\cite{extrinsic2004} & 50 & \revised{(91.03, -45.25, 0.30)} & \revised{(0.48, -6.37, -4.54)} \\
		Our w/o noise  & 1 & \revised{(90.23, -45.23, 0.27)} & \revised{(0.81, -6.66, -4.58)} \\ 
		\revised{Our w $\mathcal N_0$} & \revised{1} & \revised{(90.8, -44.83, 0.34)} & \revised{(1.45, -5.80, -4.53)}\\
		\revised{Our w $\mathcal N_1$} & \revised{1} & \revised{(90.17, -45.76, 0.71)} & \revised{(2.16, -5.00, -5.56)}\\
		\bottomrule
	\end{tabular}
	\label{tab:experiment_lidar_cam_accu}
\end{table}  

\subsubsection{camera-LiDAR extrinsic calibration}
\label{sec:exp_cam_lidar}

The extrinsic calibration between a 3D LiDAR and a camera is conducted on the mobile robot shown in Fig.~\ref{fig:mobile_robot} using one camera and one LiDAR. The extrinsic calibration results based on the proposed method is also compared with the results based on the algorithm~\cite{extrinsic2004}, which can be extended to calibrate the extrinsics between the 3D LiDAR and the camera. The extrinsic calibration results are presented in TABLE~\ref{tab:experiment_lidar_cam_accu}. The results show that the calibration accuracy of the proposed method with only single scan is comparable with the extended algorithm~\cite{extrinsic2004}.

\subsubsection{LiDAR-LiDAR extrinsic calibration}
\label{sec:exp_lidar_lidar}

The performance of LiDAR-LiDAR extrinsic calibration is evaluated using the two LiDARs on the backpack scanner as shown in Fig.~\ref{fig:backpack}: one LiDAR is mounted on the top and the other is tilted and placed at the back. \revised{TABLE~\ref{tab:experiment_lidar_lidar_accu} shows the comparison of the translation estimations between our method and ~\cite{jeong2018complex}. The results show that the algorithm in~\cite{jeong2018complex} does not work well because the two LiDARs have few overlapped FoV area, which has no impact on the proposed method.}

\begin{table}[tb]
	\centering
	\caption{LiDAR-LiDAR extrinsic calibration results.}
	\begin{tabular}{c c c } 
		\toprule
		Method & Frames & Translation vector (cm)  \\ 
		\midrule
		GT & - & \revised{(-22.55, 13.02, -34.92)} \\ 
		\revised{Jeong~\cite{jeong2018complex}} & 1 & \revised{(-20.52, 11.43, -35.01)}\\
		Our w/o noise & 1 & \revised{(-22.59, 12.48, -33.86)} \\ 
		\revised{Our w $\mathcal N_0$} & \revised{1} & \revised{(-22.30, 12.39, -33.96)} \\
		\revised{Our w $\mathcal N_1$} & \revised{1} & \revised{(-20.02, 12.38, -35.07)} \\
		\bottomrule
	\end{tabular}
	\label{tab:experiment_lidar_lidar_accu}
\end{table}

\subsection{\revised{Multi-sensor calibration accuracy}}
\label{sec:exp_multisensor_calib}

\revised{We design a series of experiments to qualitatively and quantitatively analyze the case of multi-sensor calibration accuracy. For validation, we collect different data including three platforms that equipped with various sensor suite. These platforms are referred as the TLS (Fig.~\ref{fig:kaleido}), the backpack scanner (Fig.~\ref{fig:backpack}) and the mobile robot (Fig.~\ref{fig:mobile_robot}), respectively. For each platform, the calibration procedure is repeated ten times and the average results are calculated for a robust analysis. 
	Since full CAD drawings of three platforms are provided, we compare the rotation and translation of the calibration result $T^{x}_{y}$ to the CAD drawings, which is treated as GT. The rotation is represented by Euler angles of the XYZ order, the translation is expressed in the three directions of XYZ.}

\subsubsection{TLS}
In terms of the TLS, the 3D LiDAR scan is derived from 2D LiDAR rotated by a servo motor, which has a FoV of $360^\circ$ with angle resolution of $0.13^\circ$. So the stitched 3D LiDAR scan is panoramic point cloud of the storeroom. The top camera is named as camera0, and the other sensors are named accordingly: camera1, LiDAR0, \textit{etc}. The results of the relative poses are shown in TABLE~\ref{tab:experiment_kaleido}. 

\begin{table}[tb]
	\centering
	\caption{Extrinsic calibration results of TLS.}
	\begin{tabular}{c 
			>{\centering\arraybackslash}m{0.5cm} 
			>{\centering\arraybackslash}m{2.4cm} 
			>{\centering\arraybackslash}m{2.4cm}} 
		\toprule  
		\revised{Object} & \revised{Method} & \revised{Rotation (deg)} & \revised{Translation (cm)} \\
		\midrule 
		\revised{$T^{C1}_{C0}$} & \revised{GT our} & \revised{(28.00, 0.00, 0.00) (27.69, 0.27, 0.29)} & \revised{(0.00, 2.80, 0.00) (0.04, 2.75, 0.61)}\\
		\revised{$T^{L0}_{C0}$} & \revised{GT our} & \revised{(-14.00, 0.00, 0.00) (-13.87, 0.79, 0.05)} & \revised{(0.00, -1.40, 2.00) (0.02, -1.96, 2.08)} \\
		\bottomrule
	\end{tabular}
	\label{tab:experiment_kaleido}
\end{table}

\subsubsection{backpack scanner}
\begin{figure}
	\centering
	\begin{subfigure}{0.24\textwidth}
		\includegraphics[width=\textwidth]{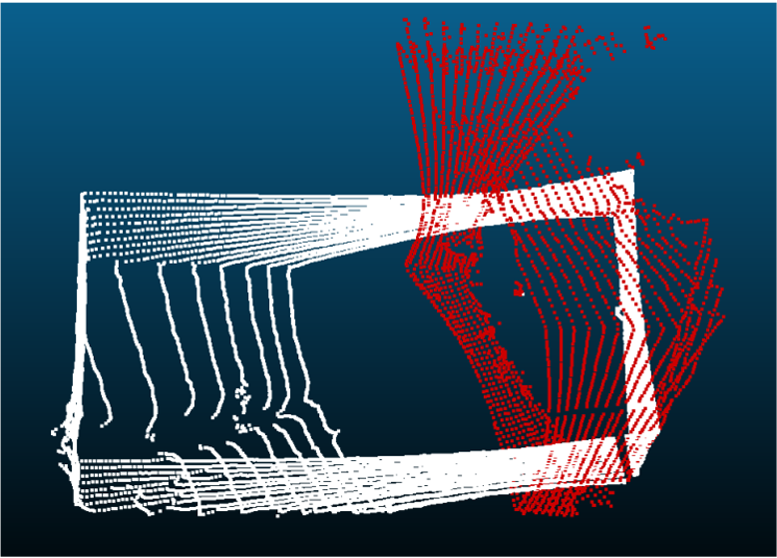}
		\caption{}
		\label{fig:vis_res1}
	\end{subfigure}\hfil
	\begin{subfigure}{0.24\textwidth}
		\includegraphics[width=\textwidth]{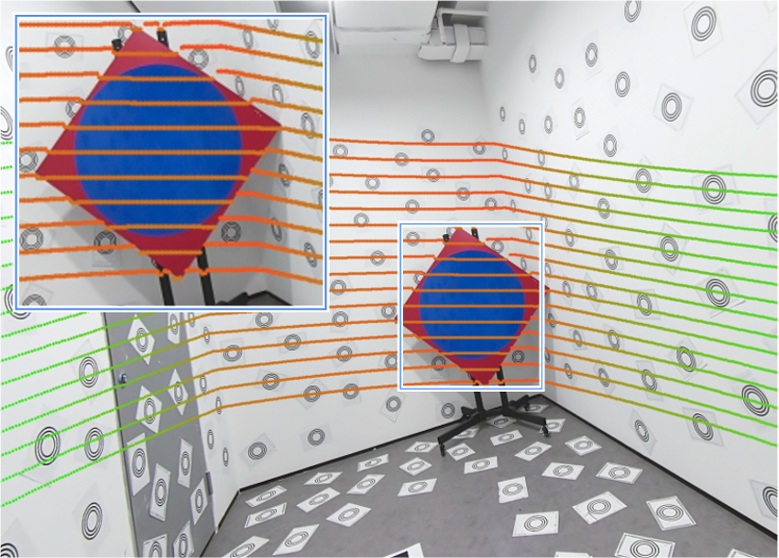}
		\caption{}
		\label{fig:vis_res2}
	\end{subfigure}\hfil
	\caption{ \revised{Extrinsic calibration result of backpack scanner: (a) Two LiDARs' scans are registered according to $T^{L1}_{L0}$. (b) LiDAR scan is projected onto the image according to $T^{C0}_{L0}$. The color of the points vary representing the distance.}}
	\label{fig:vis_result}
\end{figure}

Regarding of the backpack scanner, one of the four fisheye cameras is chosen to be camera0, and the top LiDAR is chosen to be LiDAR0. 
\revised{The results of relative poses among cameras and LiDARs} are shown in TABLE~\ref{tab:experiment_backpack}. 
\revised{In Fig.~\ref{fig:vis_res1}, two LiDAR scans are registered according to extrinsics between LiDAR0 and LiDAR1, the white one denotes LiDAR0 and the red one denotes LiDAR1. In Fig.~\ref{fig:vis_res2}, a LiDAR scan is projected onto the image using the transform from LiDAR0 to camera0, which offers a visual evidence of correspondence between the boundaries of object in the point cloud and the edges of object in the image. The color of the points varying from green to red represents the distance from near to far.}

\revised{The error of LiDAR0-LiDAR1 and camera0-LiDAR1 calibration are notably larger compared to the others, because the LiDARs have range accuracy of $2cm$.} 

\begin{table}[tb]
	\centering
	\caption{Extrinsic calibration results of backpack scanner. }
	\begin{tabular}{c 
			>{\centering\arraybackslash}m{0.5cm} 
			>{\centering\arraybackslash}m{2.6cm} 
			>{\centering\arraybackslash}m{2.6cm}}
		\toprule  
		\revised{Object} & \revised{Method} & \revised{Rotation (deg)} & \revised{Translation (cm)} \\
		\midrule 
		\revised{$T^{C1}_{C0}$} & \revised{GT our} & \revised{(0.00, -90.00, 0.00) (0.18, -90.16, 0.49)} & \revised{(-3.12, 0.0, -2.12) (-2.81, 0.01, -3.09)} \\
		\revised{$T^{C2}_{C0}$} & \revised{GT our} & \revised{(0.00, -180.0, 0.00) (0.09, -179.82, 0.26)} & \revised{(0.00, 0.00, -6.23) (0.26, 0.02, -5.88)} \\
		\revised{$T^{C3}_{C0}$} & \revised{GT our} & \revised{(0.00, 90.00, 0.00) (0.23, 89.78, 0.11)} & \revised{(3.12, 0.00, -3.12) (3.07, 0.01, -2.82)}\\
		\revised{$T^{L0}_{C0}$} & \revised{GT our} & \revised{(0.00, -15.00, 90.00) (-1.34, -14.66, 88.12)} & \revised{(61.38, 0.00, -3.81) (63.57, 0.30, -5.03)}\\
		\revised{$T^{L1}_{L0}$} & \revised{GT our} & \revised{(0.00, -75.00, -30.00) (1.41, -75.58, -31.39)} & \revised{(-22.55, 13.02, -34.92) (-21.64, 12.62, -34.28)}\\
		\bottomrule
	\end{tabular}
	\label{tab:experiment_backpack}
\end{table}

\subsubsection{mobile robot}
Regarding of the mobile robot, one of the four wide-angle cameras is chosen to be camera0, and the \revised{calibration results between any two sensors are shown in TABLE~\ref{tab:experiment_antman}}. 

\begin{table}[tb]
	\centering
	\caption{Extrinsic calibration results of mobile robot.}
	\begin{tabular}{c 
			>{\centering\arraybackslash}m{0.5cm} 
			>{\centering\arraybackslash}m{2.6cm} 
			>{\centering\arraybackslash}m{2.6cm}}
		\toprule  
		\revised{Object} & \revised{Method} & \revised{Rotation (deg)} & \revised{Translation (cm)} \\
		\midrule 
		\revised{$T^{C1}_{C0}$} & \revised{GT our} & \revised{(0.00, 90.00, 0.00) (0.29, 90.36, 0.43)} & \revised{(5.65, 0.00, 5.65) (5.61, 0.03, 5.36)}\\
		\revised{$T^{C2}_{C0}$} & \revised{GT our} & \revised{(0.00, 180.0, 0.00) (0.32, 179.86, 0.05)} & \revised{(0.00, 0.00, -11.29) (0.24, 0.06, -11.02)}\\
		\revised{$T^{C3}_{C0}$} & \revised{GT our} & \revised{(0.00, -90.00, 0.00) (0.30, -89.41, 0.11)} & \revised{(-5.65, 0.00, -5.65) (-5.42, 0.03, -5.42)}\\
		\revised{$T^{L0}_{C0}$} & \revised{GT our} & \revised{(90.00, -45.00, 0.00) (89.77, -45.18, 0.30)} & \revised{(0.00, -5.93, -5.49) (0.63, -6.65, -4.93)}\\
		\bottomrule
	\end{tabular}
	\label{tab:experiment_antman}
\end{table}

Because of the accurately reconstructed panoramic 3D points of the storeroom and the robust localization method, \revised{most extrinsics in the above experiments match well with the GT, which is eligible to verify the accuracy and robustness of the proposed method.}

\subsection{\ds{Marker comparison}}
\label{sec:exp_marker_comparison}
\revised{For marker selection, we conduct another experiment to compare the calibration accuracy of the CCTags with the AprilTags. To be specific,  the  AprilTags  and  the  CCTags are  used  to  localize  the  left  and  the right  camera  of  the stereo camera with  single frame respectively, then the extrinsics can be derived. 
	The comparison results of the extrinsics are presented in TABLE~\ref{tab:stereo_calib_result_comparison}, which shows that the CCTags outperform the AprilTags in terms of calibration accuracy slightly.}   

\begin{figure}[t]
	\centering
	\begin{subfigure}{0.2\textwidth}
		\includegraphics[width=\textwidth]{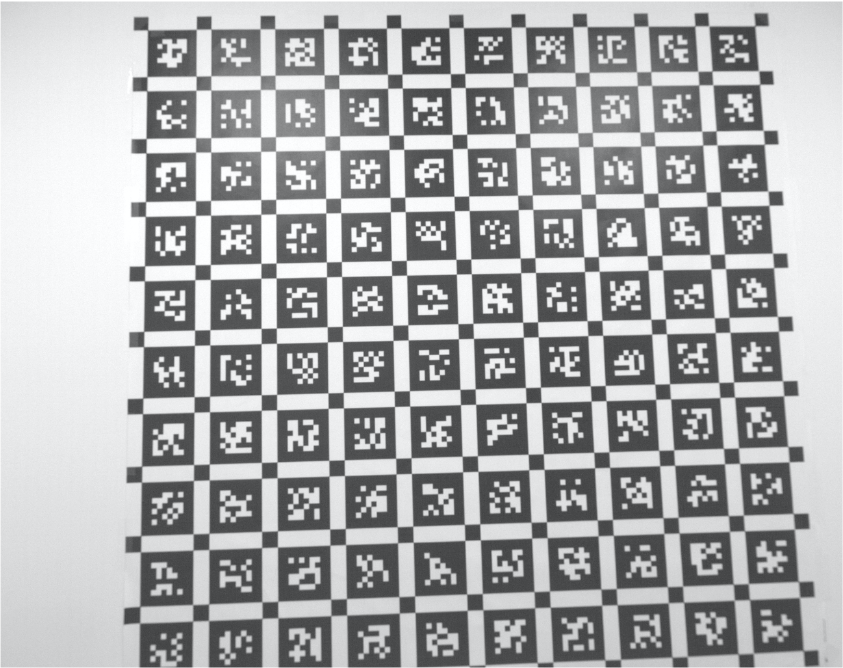}
		\caption{}
		\label{fig:undist_apriltags}
	\end{subfigure}\hfil
	\begin{subfigure}{0.2\textwidth}
		\includegraphics[width=\textwidth]{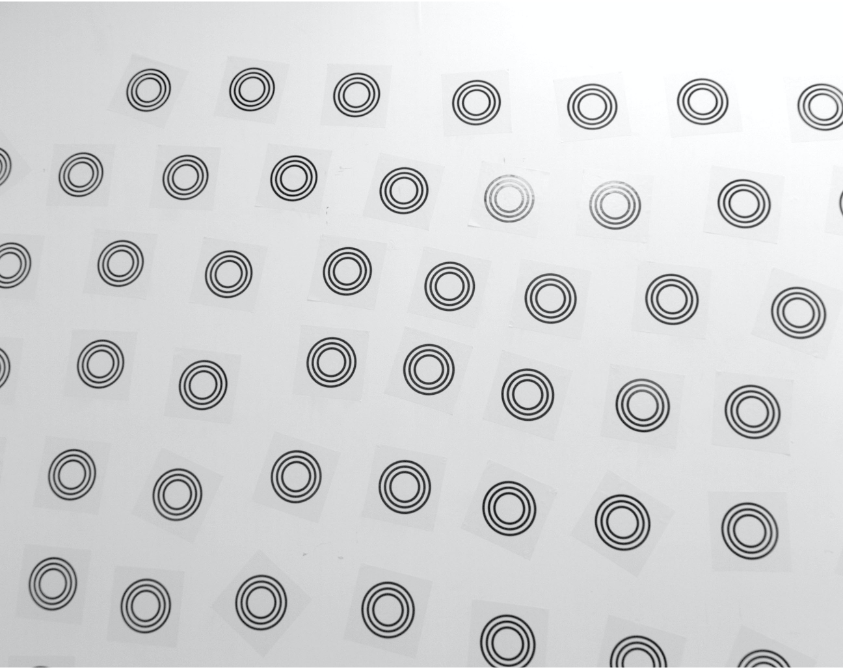}
		\caption{}
		\label{fig:undist_cctags}
	\end{subfigure}\hfil
	\caption{\revised{Markers captured from a distance of 2m: (a) The captured 100 AprilTags. (b) The captured 40 CCTags.}}
	\label{fig:undist_result}
\end{figure}

\begin{table}
	\centering
	\caption{\revised{Comparison results of the stereo camera}}
	\begin{tabular}{>{\centering\arraybackslash}m{1cm}  >{\centering\arraybackslash}m{1cm}  >{\centering\arraybackslash}m{1.5cm}
			>{\centering\arraybackslash}m{1.5cm}
			>{\centering\arraybackslash}m{1.5cm}} 
		\toprule
		\revised{Method} & \revised{Frames} & \revised{Baseline (cm)} & \revised{Reprojection Error (px)} & \revised{Standard Deviation (cm)}\\ 
		\midrule
		\revised{Kalibr~\cite{kalibr}} & \revised{150} & \revised{12.09} & \revised{0.18} & \revised{0.01} \\ 
		\revised{AprilTags} & \revised{1} & \revised{12.01} & \revised{0.62} & \revised{0.13} \\
		\revised{CCTags} & \revised{1} & \revised{12.11} & \revised{0.31} & \revised{0.04}\\
		\bottomrule
	\end{tabular}
	\label{tab:stereo_calib_result_comparison}
\end{table}

\section{Conclusion}
\label{sec:conclusion}

In this paper, we seek for an efficient calibration solution, which should involve few operations, require no expertise, and adapt to various configurations of multiple cameras and LiDARs. 
For this purpose, we proposed a single-shot calibration approach based on a panoramic infrastructure. This reference environment is carefully designed so that a camera or a LiDAR can be robustly localized, from which the relative transformations of sensors can be derived.
We also proposed an economic method for reconstructing the infrastructure using low-end stereo camera, which achieves comparable accuracy with expensive professional 3D scanners, e.g., Leica BLK360. 

\subsection{Limitations}
The localization of a LiDAR solely relies on the ICP registration results between the frame of 3D scans and the infrastructure. In case a LiDAR scan contains no salient geometric features, the ICP matching might fail. 
In order to uniquely localize LiDARs with narrow FoV, two remedial measures can be adopted. 
First, multiple LiDARs ought be calibrated by pairwise approaches, then their scans assemble a larger scan, which is more likely to be correctly localized in the infrastructure.
Second, the infrastructure room should be enriched with geometric features by, for instance, increasing the number of planes. 

\subsection{Future work}
We would like to extend our panoramic infrastructure for calibrating other types of sensors, for example, camera-IMU (Inertial Measurement Unit) configuration. Moreover, intrinsic parameters of cameras could also be calibrated together with extrinsic transformations using the proposed infrastructure. 
We also want to check out whether it is beneficial to optimize extrinsic parameters of cameras and LiDARs in a united formulation.

	\bibliographystyle{IEEEtran}
	\bibliography{calibration.bib}
\end{document}